%% file: neurips_2025.tex
\documentclass{article}

\usepackage[preprint]{neurips_2025}

\usepackage[utf8]{inputenc} 
\usepackage[T1]{fontenc}    
\usepackage{hyperref}       
\usepackage{url}            
\usepackage{booktabs}       
\usepackage{amsfonts}       
\usepackage{nicefrac}       
\usepackage{microtype}      
\usepackage{graphicx}
\usepackage{multirow} 
\usepackage{enumitem}
\usepackage{tabulary}
\usepackage{colortbl}
\usepackage{wrapfig}
\usepackage{amsmath, amssymb}
\usepackage{algorithm}
\usepackage{algpseudocodex}
\usepackage{hyperref}
\usepackage{booktabs}

\algrenewcommand\algorithmiccomment[1]{\hfill{\scriptsize\textcolor{gray}{// #1}}}

\title{Active Test-time Vision-Language Navigation}

\author{%
  Heeju Ko\textsuperscript{\normalfont 1} \quad
  Sungjune Kim\textsuperscript{\normalfont 1} \quad
  Gyeongrok Oh\textsuperscript{\normalfont 1} \quad
  Jeongyoon Yoon\textsuperscript{\normalfont 1} \\
  \textbf{Honglak Lee}\textsuperscript{\normalfont 2} \quad
  \textbf{Sujin Jang}\textsuperscript{\normalfont 3} \quad
  \textbf{Seungryong Kim}\textsuperscript{\normalfont 4} \quad
  \textbf{Sangpil Kim}\textsuperscript{\normalfont 1} \quad\\ \\
  \textsuperscript{\normalfont 1} Korea University \quad \textsuperscript{\normalfont 2} University of Michigan\quad
  \textsuperscript{\normalfont 3} Samsung AI Center, DS Division\quad\\
  \textsuperscript{\normalfont 4} Korea Advanced Institute of Science \& Technology\\
}

\begin{document}

\maketitle

\input{sec/0_abstract}

\input{sec/1_introduction}

\input{sec/2_related}

\input{sec/3_method}

\input{sec/5_exp_main}
\input{sec/6_conclusion}

\bibliographystyle{unsrt}
\bibliography{ref}

\newpage
\input{sec/7_appendix}

\end{document}

%% file: sec/0_abstract.tex
\begin{abstract}

Vision-Language Navigation~(VLN) policies trained on offline datasets often exhibit degraded task performance when deployed in unfamiliar navigation environments at test time, where agents are typically evaluated without access to external interaction or feedback.
Entropy minimization has emerged as a practical solution for reducing prediction uncertainty at test time; however, it can suffer from accumulated errors, as agents may become overconfident in incorrect actions without sufficient contextual grounding.
To tackle these challenges, we introduce \textsc{ATENA} (Active TEst-time Navigation Agent), a test-time active learning framework that enables a practical human-robot interaction via episodic feedback on uncertain navigation outcomes.
In particular, \textsc{ATENA} learns to increase certainty in successful episodes and decrease it in failed ones, improving uncertainty calibration.
Here, we propose \textit{mixture entropy optimization}, where entropy is obtained from a combination of the action and pseudo-expert distributions—a hypothetical action distribution assuming the agent's selected action to be optimal—controlling both prediction confidence and action preference. In addition, we propose a \textit{self-active learning} strategy that enables an agent to evaluate its navigation outcomes based on confident predictions.
As a result, the agent stays actively engaged throughout all iterations, leading to well-grounded and adaptive decision-making.
Extensive evaluations on challenging VLN benchmarks—REVERIE, R2R, and R2R-CE—demonstrate that \textsc{ATENA} successfully overcomes distributional shifts at test time, outperforming the compared baseline methods across various settings.

\end{abstract}

%% file: sec/1_introduction.tex
\section{Introduction}
Vision-Language Navigation (VLN) is a fundamental multimodal task in embodied AI systems, which requires an agent to interpret natural language instructions and navigate through complex visual environments~\cite{anderson2018vision}.
Despite recent advancements in VLN, distributional shifts between offline training and online testing environments remain a critical challenge for robust and reliable deployment~\cite{parvaneh2020counterfactual, li2022envedit}.
To address this issue, many prior works focus on enhancing generalizability during offline training to better handle potential domain shifts~\cite{ chen2022think, chen2021history, an2024etpnav}.
However, these approaches are limited in addressing real-world variability, as collecting diverse expert demonstrations across environments is often impractical.
Therefore, test-time adaptation (TTA)—the ability to directly adapt to test-time environments—is crucial for real-world robotic navigation.

Test-Time Adaptation (TTA) refines a pre-trained model during inference using unsupervised signals—such as prediction entropy~\cite{wang2020tent}, consistency~\cite{Wang_2022_CVPR}, or pseudo-labels~\cite{goyal2022test}—offering a practical yet challenging approach to improving robustness at test time.
Entropy minimization is one of the widely accepted TTA strategies, based on the assumption that greater model certainty correlates with improved accuracy during inference~\cite{wang2020tent, yang2024towards, zhou2025test}.
However, applying entropy minimization uniformly across all decision points in sequential tasks such as VLN may cause the policy to overfit to failure patterns, thereby increasing the likelihood of incorrect actions.
Consequently, the policy accumulates errors throughout iterations and loses resilience on failure cases.
Thus, blindly increasing prediction certainty without considering the navigation status leads to suboptimal behaviors.

How, then, can we provide the contextual cues necessary for a VLN agent to properly leverage entropy as a meaningful signal at test time?
To answer this, we propose an active learning (AL)~\cite{settles2009active, li2024survey} strategy, enabling the agent to query a human oracle for necessary contextual labels.
Here, we must consider practical constraints in the online test-time navigation setting for utilizing human feedback: (1) Latency—human involvement should not introduce delays during navigation rollout; and (2) Accessibility—human input must be intuitive, requiring minimal expertise and effort. Therefore, it is unrealistic to expect human feedback at the same level of detail as stepwise expert demonstrations in VLN at test time.

To address these practical concerns, we define the active label as an episodic, binary evaluation indicating navigation success or failure, rather than requiring detailed stepwise supervision. 
Inspired by the uncertainty sampling paradigm in AL~\cite{wu2022entropy, tamkin2022active, tifrea2023margin}, we design the agents to inquire feedback whenever the average action uncertainty throughout each navigation task exceeds a predefined threshold. 
Given the sparsity of the feedback, we introduce a novel technique called \textit{mixture entropy optimization}~(MEO) to effectively leverage it. Specifically, based on the binary outcome, we guide entropy optimization by minimizing entropy for successful navigation and maximizing it for failed ones. Here, entropy is derived from a mixture of two distributions: an action distribution, representing the likelihood assigned to each possible action, and a pseudo-expert distribution, a one-hot probability distribution centered on the agent's chosen action, assuming this action as optimal. By combining these two distributions, MEO not only controls the certainty of the decisions but also explicitly suppresses incorrect actions and encourages actions that led to successful navigation.

Additionally, we introduce a novel paradigm of \textit{self-active learning}~(SAL), enabling the navigation agent to remain actively engaged throughout all iterations for continuous feedback. Traditional AL methods typically request human feedback only when the model prediction is uncertain, potentially overlooking the subtle errors that are hidden beneath high confidence in certain predictions. In contrast, our method allows the agent to determine the navigation outcome by itself in relatively certain predictions. This is achieved through a self-prediction head, initialized at test time and trained during streaming test episodes using both human-provided labels and the agent’s own predicted outcomes. As a result, the agent receives continuous guidance for the direction of entropy optimization, which is crucial for precise adaptation. Ultimately, SAL reduces reliance on human intervention, thereby improving the agent’s autonomy and robustness in online test-time environments.

We name our overall framework as \textsc{ATENA}~(Active TEst-time Navigation Agent), and validate its effectiveness through comprehensive evaluation on challenging VLN benchmarks: REVERIE~\cite{qi2020reverie}, R2R~\cite{anderson2018vision}, and R2R-CE~\cite{krantz2020beyond}. \textsc{ATENA} achieves significant gains over the underlying target policies and outperforms strong TTA baselines. Our empirical results and in-depth analysis indicate that \textsc{ATENA} effectively addresses test-time distribution shifts and provides a strong foundation for future research on active human-robot interaction in vision-and-language navigation.

The contributions of this work are summarized as follows:
\begin{itemize}[leftmargin=*]
    \item We introduce \textsc{ATENA}, the first active learning framework for online VLN that leverages human input to guide entropy-based optimization.
    \item Mixture entropy optimization enhances confidence calibration by explicitly suppressing incorrect actions and encouraging desired actions.
    \item The self-active learning phase provides a strategic solution to provide continuous active labels, with reduced burden of human labeling in online environment.
\end{itemize}

%% file: sec/2_related.tex
\section{Related Work}
\label{gen_inst}
\subsection{Vision-Language Navigation}
\label{sec2-1:vln}
Vision-Language Navigation (VLN) is a pivotal task of
bridging human communications with embodied AI system~\cite{gao2024vision,wu2024vision,gu2022vision}. The sequential natures of the decision making process in VLN led early research to adopt recurrent neural network-based architectures~\cite{anderson2018vision,an2021neighbor,wang2020vision}. Following works utilized the Transformer network~\cite{vaswani2017attention} to capture complex multimodal dependencies and achieved substantial performance gains~\cite{chen2021history, chen2022think, liu2024volumetric,hao2020towards,wang2023dual, landi2021multimodal, li2019robust}. However, these offline training methods merely anticipate domain shifts and suffer a performance degradation when the online navigation environment deviates from the training distribution. To overcome the discrepancy, large-language models came into play as zero-shot navigation agents, but their reasoning capabilities without fine-tuning have yet to yield reliable performances~\cite{zhou2024navgpt,long2024discuss,zhou2024navgpt2}. A recent approach focuses on directly adapting an offline-trained policy to online test time environment using unsupervised entropy minimization~\cite{gao2024fast}. In this work, we explore how the core principle of VLN—\textit{human-robot interaction}—can be effectively leveraged to facilitate online test-time adaptation.

\vspace{-0.1em}
\subsection{Entropy-based Test-time Adaptation}
\label{sec2-2:tta}
Entropy minimization is a widely adopted learning objective in domain adaptation~\cite{ma2022context, vu2019advent, wu2021entropy} and semi-supervised learning~\cite{grandvalet2004semi,berthelot2019mixmatch,wu2021semi}. Recently, entropy minimization has emerged as a foundational technique in test-time adaptation (TTA) due to its simplicity and effectiveness in the absence of labeled target data.~\cite{liang2025comprehensive, xiao2024beyond}. The core idea is to encourage the model to make confident predictions by minimizing the entropy of output distributions at test time, assuming that well-adapted models should be confident on in-distribution samples. Tent~\cite{wang2020tent} introduced a lightweight yet effective approach that minimizes prediction entropy by updating only batch normalization parameters during test time. Building upon this, numerous studies across various fields began integrating entropy minimization into their TTA strategies~\cite{gao2024unified, gong2022note, yang2024towards, zhou2025test}. In VLN, FSTTA~\cite{gao2024fast} extends entropy minimization by accounting for the sequential and episodic nature of the task. Although effective in many settings, blindly minimizing the entropy can lead to the propagation of overconfident mistakes, making it particularly problematic in sequential tasks like VLN where early errors can cascade~\cite{tan2025uncertainty,lee2024entropy}. This work addresses the problem by enabling agents to actively query oracles for feedback on uncertain navigation outcomes, providing crucial guidance for entropy-based test-time adaptation.

\vspace{-0.1em}
\subsection{Active Learning}
\label{sec2-3:active}

Active Learning (AL) is a machine learning strategy designed to efficiently reduce labeling costs by selectively querying labels for the most uncertain or informative data points~\cite{settles2009active, ren2021survey, budd2021survey, li2024survey}. Traditional AL methods primarily utilize uncertainty sampling, prioritizing data points when the model's predictive confidence is low; typical metrics for quantifying uncertainty include entropy, margin sampling, and least-confidence measures~\cite{wu2022entropy, tamkin2022active, tifrea2023margin}. Initially focused on relatively simple classification tasks, AL techniques have progressively evolved to tackle increasingly complex, real-world scenarios~\cite{mundt2023wholistic, ayub2022few, park2025active, fu2021transferable}. Recent advancements further extend AL concepts into TTA, enabling models to dynamically adapt during inference by utilizing uncertainty estimates, thus reducing the reliance on extensive retraining or large amounts of labeled data~\cite{gui2024active, wang2025effortless}. Inspired by these developments, our research pioneers the integration of Active Test-Time Adaptation into VLN, effectively overcoming practical constraints in real-world navigation.

%% file: sec/3_method.tex
\section{Method}

\begin{figure}[t]
\begin{center}
\includegraphics[width=\linewidth]{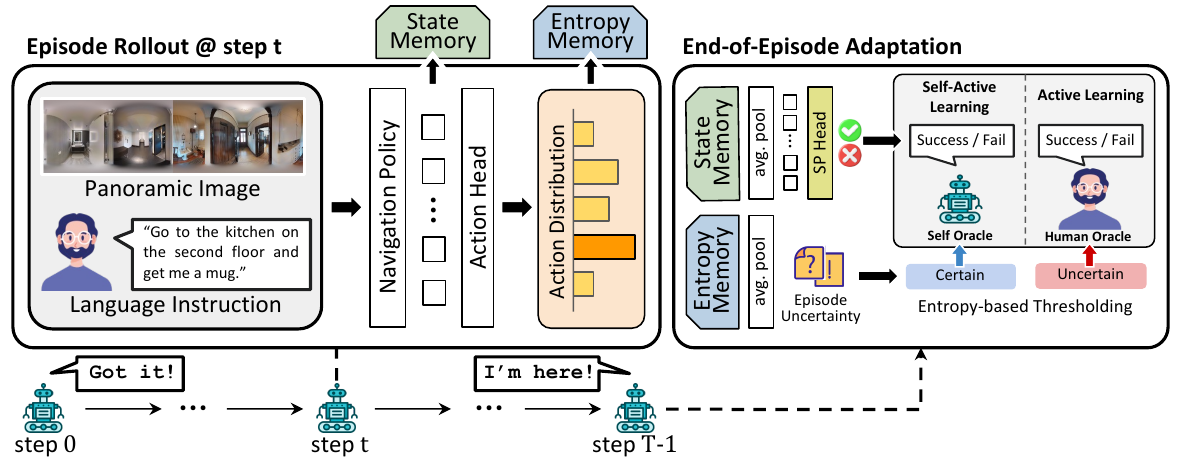}
\end{center}
\vspace{-1.0em}
\caption{\textbf{Overview of the ATENA adaptation framework.} At each navigation step, the agent stores state and entropy information in its memory. Once the episode ends, the stored entropy is used to determine the feedback source: human oracle for uncertain episodes, and self oracle for certain episodes. Self oracle utilizes a self-prediction head, trained during online test-time, enabling the agent to autonomously predict navigation success or failure by itself.}
\label{fig:selective}
\end{figure}

\subsection{Task Description}

Vision-Language Navigation (VLN) tasks an agent with interpreting natural language instructions $I$ to navigate through a visual environment. Starting from an initial visual observation $o_0$, at each timestep $t$, the agent perceives a visual observation $o_t$, selects an action $a_t$ according to its policy $\pi_{\theta}$, and transitions into the next state. Repeating this process until the agent selects a stopping action produces a trajectory $\tau = \{(o_t, a_t)\}_{t=0}^{T-1}$, where $T$ is the total number of steps taken. In this work, we specifically consider an online VLN scenario where the navigation policy encounters a stream of test instructions and environments during deployment.

\subsection{Overview}

Our proposed framework, \textsc{ATENA} (Active TEst-time Navigation Agent), enables active learning in online VLN by integrating human guidance into entropy-based optimization. \textsc{ATENA} consists of two core components:

\begin{itemize}[leftmargin=*]
\item \textbf{Mixture Entropy Optimization (MEO):} A method that refines the agent's policy using outcome-conditioned entropy signals, leveraging a pseudo-expert-guided action distribution to more effectively amplify correct behavior and penalize failure.

\item \textbf{Self-Active Learning (SAL):} A strategy that enables the agent to autonomously request or replace feedback based on internal uncertainty and self-assessment, allowing robust adaptation even when explicit feedback is sparse or unavailable.
\end{itemize}

Together, these solutions allow the agent to adapt online by jointly leveraging episodic outcomes and self-predicted performance, without relying on ground-truth trajectories or dense human supervision.

\subsection{Mixture Entropy Optimization (MEO)}

A traditional entropy minimization method in VLN~\cite{gao2024fast} aim to reduce uncertainty by decreasing the entropy of the predicted action distribution. However, indiscriminately minimizing entropy can reinforce confidence even in incorrect actions, leading to compounding errors during navigation. To address this, we optimize the entropy based on success or failure of the navigation episode. Specifically, we minimize it for successful episodes to reinforce the selected action, and maximize it for failed episodes to penalize incorrect decisions. Furthermore, this entropy-based test-time adaptation is facilitated by our novel mixture entropy optimization.

\subsubsection{Mixture Action Distribution}

First, we define the \textit{Mixture Action Distribution} as a convex combination of the predicted action distribution $\pi_\theta$ and a pseudo-expert distribution $q_{\text{pseudo}}$. 
The pseudo-expert distribution is a one-hot probability distribution that assigns full probability~(\textit{i.e.} 1.0) to the selected action $a_t^{\text{sel}}$, which refers to the action with the highest predicted probability under the current policy, i.e., $a_t^{\text{sel}} = \arg\max_a \pi_\theta(a \mid o_t, I)$. In other words, the pseudo-expert distribution treats as if $a_t^{\text{sel}}$ is the optimal expert action.
The mixture action distribution is formalized as:
\begin{equation}
\label{eq:mix_action}
q_{\text{mix}}(a \mid o_t, I) = \lambda q_{\text{pseudo}}(a \mid a_t^{\text{sel}}) + (1 - \lambda)\pi_\theta(a \mid o_t, I),\quad 0 \leq \lambda \leq 1.
\end{equation}

This mixture formulation sharpens the distribution around the selected action, with the combination weight $\lambda$ controlling how strongly the pseudo-expert guides the distribution. Accordingly, the entropy of the mixture action distribution at timestep $t$ is defined as:
\begin{equation}
\label{eq:mix_dist}
\mathcal{H}(q_{\text{mix}}(\cdot \mid o_t, I)) = -\sum_{a \in \mathcal{A}_{t}} q_{\text{mix}}(a \mid o_t, I)\log q_{\text{mix}}(a \mid o_t, I),
\end{equation}
where $\mathcal{A}_{t}$ is the set of all possible actions at step $t$. We average the entropy over all steps and obtain $\mathcal{H}^{\prime}(q_{\text{mix}})$ as the optimization signal of the episode. Then, the mixture entropy loss function can be formulated as:
\begin{equation}
\label{eq:mix_loss}
\mathcal{L}_{\text{mix}}=\mathbb{I}_{\text{success}}\cdot \mathcal{H}^{\prime}(q_{\text{mix}}) - (1-\mathbb{I}_{\text{success}})\cdot \mathcal{H}^{\prime}(q_{\text{mix}}),
\end{equation}
where $\mathbb{I}_{\text{success}}$ is a binary indicator that is 1 if the navigation was successful, 0 otherwise.

\subsubsection{Effect on Policy Adaptation}
\begin{figure}[t]
\begin{center}
\includegraphics[width=\linewidth]{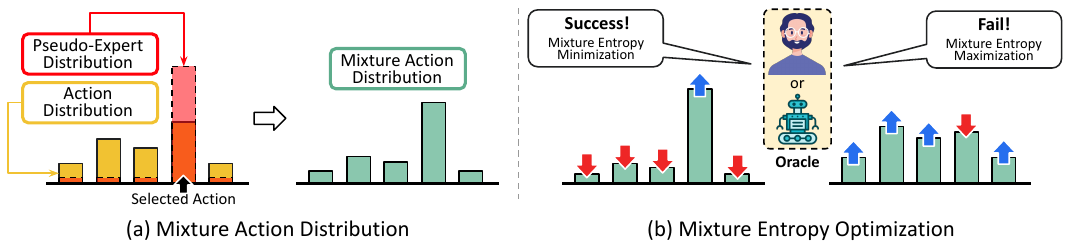}
\end{center}
\vspace{-1.0em}
\caption{\textbf{llustration of Mixture Entropy Optimization (MEO). } (a) The Mixture Action Distribution is constructed by combining the action distribution (yellow) with a pseudo-expert distribution (red). (b) Mixture entropy is minimized for successful episodes to encourage the correct actions, and maximized for failures to penalize incorrect ones.}
\label{fig:meo}
\end{figure}

Since the mixture action distribution inherently sharpens the original predicted action distribution, this strategy amplifies the feedback signal—further boosting correct actions when successful, and more strongly suppressing incorrect ones when failed (see Figure~\ref{fig:meo}). This is quantitatively evident from the selected action's probability:
\begin{equation}
\label{eq:mix_effect}
q_{\text{mix}}(a_t^{\text{sel}} \mid o_t, I) = \lambda + (1 - \lambda)\pi_\theta(a_t^{\text{sel}} \mid o_t, I),
\end{equation}

which is strictly greater than $\pi_\theta(a_t^{\text{sel}} \mid o_t, I)$ when $\lambda > 0$. As a result, entropy-based optimization applied to $q_{\text{mix}}$ exerts stronger influence on the selected action compared to directly using $\pi_\theta$. Specifically, when minimizing this entropy in successful episodes, the gradient increases $q_{\text{mix}}(a_t^{\text{sel}})$ more sharply than optimizing $\pi_\theta$ alone would. Conversely, maximizing entropy in failed episodes suppresses $q_{\text{mix}}(a_t^{\text{sel}})$ more aggressively, allowing MEO to drive stronger and more directional updates to the policy, improving sample efficiency and reducing the reliance on active learning during test time as demonstrated in Table~\ref{tab:tta_active_learning}.

\subsection{Self-Active Learning (SAL)}

Mixture Entropy Optimization enables policy refinement based on navigation outcomes, requiring episodic feedback during test-time adaptation. In practice, however, acquiring feedback—especially from human annotators—can be costly or delayed. Moreover, uncertainty alone may fail to capture subtle but critical navigation errors in seemingly confident predictions. To address these challenges, we propose \textit{Self-Active Learning}~(SAL), where the agent selectively queries human feedback or uses its own predictions on navigation outcomes to self-supervise, allowing for more robust and autonomous adaptation.

\subsubsection{Uncertainty-Guided Query Strategy}

At each timestep, the agent computes the entropy of its action distribution, $\mathcal{H}(\pi_\theta(\cdot \mid o_t, I))$, and stores it in the entropy memory. At the end of an episode, the agent determines the source of supervision $\mathcal{O}$—either Human (human-provided feedback) or Agent (self-generated feedback)—based on the average entropy. Technically, we consider $\mathcal{O}$ as a function of $\tau$ to predict $\mathbb{I}_{\text{success}}$:

\begin{equation}
\label{eq:oracle}
\mathcal{O} = 
\begin{cases}
\text{Human}, & \text{if}\;\frac{1}{T}\sum_{t=1}^{T} \mathcal{H}(\pi_\theta(\cdot \mid o_t, I)) > \delta \\[6pt]
\text{Agent}, & \text{otherwise},
\end{cases}
\end{equation}
where $\delta$ is a pre-defined uncertainty threshold. In other words, the agent requests supervision from human in uncertain navigation and self-supervise in relatively certain navigation.

\subsubsection{Self-Prediction Head}

\textbf{Predicting Navigation Outcome.} To enable autonomous self-supervision, we incorporate a self-prediction head $f_\phi$ into the pre-trained policy $\pi_{\theta}$, trained online to predict the episodic outcome (\textit{i.e.}, success or failure) from its internal states. Specifically, at step $t$, the $D$-dimensional hidden state vector $s_t \in \mathbb{R}^{D}$ is stored in the state memory and averaged over the episode as $s_{\text{avg}}$. This is then fed into the self-prediction head to determine the navigation outcome $\mathbb{I}_{\text{success}}$:
\begin{equation}
\label{eq:sal_indictor}
    \mathbb{I}_{\text{success}}=
    \begin{cases}
        1,\quad \text{if }  \sigma(f_\phi(s_{\text{avg}}))>0.5\\
        0,\quad \text{otherwise, }
    \end{cases}
\end{equation}
where $\sigma$ is the sigmoid activation function.

\textbf{Training Self-Prediction Head.} To train the self-prediction head, we use a binary cross-entropy between $f_\phi(s_{\text{avg}})$ and the binary episodic outcome $\mathbb{I}_{\text{success}} \in \{0,1\}$:
\begin{equation}
\label{eq:sal_loss}
    \mathcal{L}_{\text{self}} = -\Big[\mathbb{I}_{\text{success}} \log(\sigma(f_\phi(s_{\text{avg}})) + (1 - \mathbb{I}_{\text{success}})\log(1 - \sigma(f_\phi(s_{\text{avg}})))\Big]
\end{equation}

This loss is used during test-time adaptation regardless of the label source. If the feedback oracle is human, we assume that $\mathbb{I}_{\text{success}}$ is mostly accurate. Alternatively, if the feedback oracle is the agent itself, this can be interpreted as a self-training paradigm with pseudo label derived from the agent's own assessment of task completion, enabling continual self-improvement without external supervision as shown in Table~\ref{tab:self_active_learning}. Algorithm~\ref{algorithm:sal} summarizes the full adaptation process of SAL.

\textbf{Total Adaptation Objective of ATENA.} We combine the mixture entropy loss from Eq.~\ref{eq:mix_loss} and the self-prediction loss into a unified test-time adaptation objective:
\begin{equation}
\label{eq:total_loss}
    \mathcal{L} = \mathcal{L}_{\text{mix}} + \gamma \mathcal{L}_{\text{self}},    
\end{equation}
where $\gamma$ balances the influence of self-assessment. This joint objective reinforces correct decisions, penalizes errors, and improves the agent’s ability to assess its own performance during deployment.

\begin{algorithm}[t!]
\caption{Self-Active Learning for Online Adaptation}
\label{algorithm:sal}
\begin{algorithmic}[1]  
\Require Pre-trained policy $\pi_\theta$, entropy threshold $\delta$, learning rate $\eta$, loss weight $\gamma$
\State Initialize parameters $\theta$, $\phi$
\For{each episode}
    \State Follow instruction $I$ and collect trajectory $\tau = \{(o_t, a_t)\}_{t=0}^{T-1}$
    \State Compute average entropy $\bar{\mathcal{H}} = \frac{1}{T} \sum_{t=0}^{T-1} \mathcal{H}(\pi_\theta(\cdot \mid o_t, I))$ \hfill (Eq.~\ref{eq:oracle}) 
    \If{$\bar{\mathcal{H}} > \delta$}
        \State Receive human feedback: $\mathbb{I}_{\text{success}} \gets \mathcal{O}_{\text{Human}}(\tau)$
    \Else
        \State Receive agent feedback: $\mathbb{I}_{\text{success}} \gets \mathcal{O}_{\text{Agent}}(\tau)~ $(Eq.~\ref{eq:sal_indictor})
    \EndIf
    \State Compute mixture entropy loss $\mathcal{L}_{\text{mix}}$ \hfill (Eq.~\ref{eq:mix_loss})
    \State Compute self-prediction loss $\mathcal{L}_{\text{self}}$ \hfill (Eq.~\ref{eq:sal_loss})
    \State $\mathcal{L} = \mathcal{L}_{\text{mix}} + \gamma \mathcal{L}_{\text{self}}$ \hfill (Eq.~\ref{eq:total_loss})
    \State Update: $(\theta, \phi) \gets (\theta, \phi) - \eta \nabla_{\theta, \phi} \mathcal{L}$
\EndFor
\Return Adapted policy parameters $(\theta^*, \phi^*)$
\end{algorithmic}
\end{algorithm}

%% file: sec/5_exp_main.tex
\section{Experiments}

\vspace{-0.1em}
\subsection{Datasets \& Metrics}
\vspace{-0.2em}
We conduct experiments on three challenging VLN benchmarks—REVERIE~\cite{qi2020reverie}, R2R~\cite{anderson2018vision}, and R2R-CE~\cite{krantz2020beyond}. REVERIE evaluates agents’ ability to follow high-level, goal-oriented instructions to locate remote objects in indoor environments; a navigation episode is considered successful if the agent stops within 3 meters of the target. For REVERIE, the performance is measured with Success Rate~(SR), Oracle Success Rate~(OSR), Success penalize by Path Length~(SPL) and Remote Grounding SPL~(RGSPL). R2R, in contrast, emphasizes fine-grained instruction following, providing detailed step-by-step guidance and using the same 3-meter success criterion. R2R-CE extends R2R by replacing the discrete action space with a continuous one, increasing the difficulty of low-level control and decision-making. For R2R variants, we use Trajectory Length~(TL), Navigation Error~(NE), SR and SPL as evaluation metrics.

\vspace{-0.1em}
\subsection{Baselines}
\vspace{-0.2em}
For experiments, we apply our \textsc{ATENA} on pre-trained HAMT~\cite{chen2021history}, DUET~\cite{chen2022think} BEVBert~\cite{an2023bevbert}, ETPNav~\cite{an2024etpnav} and GOAT~\cite{wang2024vision}. HAMT is an end-to-end transformer-based VLN network trained via reinforcement learning. DUET exploits both global topology and local visual information for decision-making. BEVBert enhances spatial understanding by encoding the environment into Bird's-Eye-View representation. EPTNav emphasizes long-range planning for agents operating in continuous environments. Lastly, GOAT is a unified structural causal model for VLN. We compare our method against Tent~\cite{wang2020tent} and FSTTA~\cite{gao2024fast}. Tent is a TTA method that minimizes entropy to adjust normalization statistics. FSTTA further applies the concept of entropy minimization to the sequential VLN task. However, due to a reported issue in the official codebase~\footnote{ \scriptsize \url{https://github.com/Feliciaxyao/ICML2024-FSTTA/issues/1}}, we re-implement the method to ensure accurate evaluation. Throughout our experiments, a $\dagger$ indicates results obtained from our version.

\vspace{-0.5em}
\begin{table*} 
	\centering
	\caption{Experimental results on the REVERIE dataset. $\dagger$ implies that the results are obtained from our re-implementation~(same for Table~\ref{tab:r2r} and  Table~\ref{tab:r2rce}).}
\label{tab:reverie}
\resizebox{0.98\textwidth}{!}{
	\begin{tabular}{l|*{4}{c}|*{4}{c}|*{4}{c}}
            \toprule
            \multicolumn{1}{c|}{\multirow{2}{*}{\textbf{Methods}}}  & 
            \multicolumn{4}{c|}{\textbf{Val Seen}} & 
            \multicolumn{4}{c|}{\textbf{Val Unseen}} & 
            \multicolumn{4}{c}{\textbf{Test Unseen}}  \\
            \cmidrule(lr){2-5} \cmidrule(lr){6-9} \cmidrule(lr){10-13}
             & OSR~$\uparrow$ & SR~$\uparrow$ & SPL~$\uparrow$ & RGSPL~$\uparrow$ & OSR~$\uparrow$ & SR~$\uparrow$ & SPL~$\uparrow$ & RGSPL~$\uparrow$ & OSR~$\uparrow$ & SR~$\uparrow$ & SPL~$\uparrow$ & RGSPL~$\uparrow$ \\
		\midrule
            HAMT~\cite{chen2021history}
		& 47.65 & 43.29 & 40.19 & 25.18
		& 36.84 & 32.95 & 30.20 & 17.28
		& 33.41 & 30.40 & 26.67 & 13.08 \\
            w/ TENT$^{\dagger}$~\cite{wang2020tent}
		& 46.03 & 43.43 & 40.78 & 25.81
		& 32.60 & 30.56 & 28.23 & 14.48
		& 25.06 & 23.73 & 21.78 & 10.82 \\
		w/ FSTTA$^{\dagger}$~\cite{gao2024fast}
		& 48.21 & 42.87 & 39.56 & 24.58
		& 36.78 & 32.89 & 30.51 & 17.20
		& 33.39 & 30.39 & 26.65 & 13.61\\
            w/ \textsc{ATENA}~(\textbf{Ours})
		& \cellcolor{orange!20}52.92 & \cellcolor{orange!20}57.34 & \cellcolor{orange!20}48.08 & \cellcolor{orange!20}29.60
		& \cellcolor{orange!20}38.85 & \cellcolor{orange!20}34.00 & \cellcolor{orange!20}30.96 & \cellcolor{orange!20}17.51
		& \cellcolor{orange!20}38.19 & \cellcolor{orange!20}32.55 & \cellcolor{orange!20}28.38 & \cellcolor{orange!20}14.32 \\
            \cmidrule(l){1-13}
		DUET~\cite{chen2022think}
		& 73.86 & 71.75 & 63.94 & 51.14 
		& 51.07 & 46.98 & 33.73 & 23.03
		& 56.91 & 52.51 & 36.06 & 22.06\\
		w/ TENT
		& 73.72 & 71.89 & 64.06 & 50.41
		& 51.43 & 47.55 & 33.99 & 23.32
		& 57.12 & 52.61 & 36.17 & 22.16\\
		w/ FSTTA
		& 75.59 & 75.48 & 65.84 & 52.23
		& 56.26 & 54.15 & 36.41 & 23.56
		& 58.44 & 53.40 & 36.43 & 22.40\\
            w/ \textsc{ATENA}~(\textbf{Ours})
		& \cellcolor{orange!20}85.52 & \cellcolor{orange!20}84.33 & \cellcolor{orange!20}74.31 & \cellcolor{orange!20}59.99
		& \cellcolor{orange!20}71.88 & \cellcolor{orange!20}68.11 & \cellcolor{orange!20}45.82 & \cellcolor{orange!20}31.26
		& \cellcolor{orange!20}57.74 & \cellcolor{orange!20}54.28 & \cellcolor{orange!20}40.70 & \cellcolor{orange!20}25.01  \\
            \cmidrule(l){1-13}
		GOAT$^{\dagger}$~\cite{wang2024vision}
		& 82.36 & 80.74 & 73.44 & 58.82 
		& 57.97 & 53.82 & 37.52 & 27.00
		& 61.44 & 57.72 & 40.53 & 26.70 \\
		w/ TENT$^{\dagger}$
		& 82.43 & 80.74 & 73.47 & 58.75
		& 57.68 & 53.51 & 37.49 & 26.99
		& 62.00 & 57.28 & 39.82 & 26.97\\
		w/ FSTTA$^{\dagger}$
		& 82.36 & 80.74 & 73.42 & 58.82
		& 57.94 & 53.79 & 37.50 & 26.95
		& 62.35 & 57.52 & 39.49 & 26.82 \\
            w/ \textsc{ATENA}~(\textbf{Ours})
		& \cellcolor{orange!20}85.03 & \cellcolor{orange!20}83.35 & \cellcolor{orange!20}76.45 & \cellcolor{orange!20}61.60
		& \cellcolor{orange!20}70.29 & \cellcolor{orange!20}67.66 & \cellcolor{orange!20}53.15 & \cellcolor{orange!20}39.80
		& \cellcolor{orange!20}64.26 & \cellcolor{orange!20}62.03 & \cellcolor{orange!20}46.82 & \cellcolor{orange!20}31.54 \\
		\bottomrule
        \end{tabular}
        }
\end{table*}

\begin{table}[t]
\begin{minipage}{0.49\linewidth}
	\centering
	\caption{Experimental results on the R2R dataset.}
\label{tab:r2r}
\resizebox{\linewidth}{!}{
	\begin{tabular}{l|*{4}{c}|*{4}{c}}
		\toprule
		\multicolumn{1}{c|}{\multirow{2}{*}{\textbf{Methods}}} 
		& \multicolumn{4}{c}{\textbf{Val Seen}} 
		& \multicolumn{4}{|c}{\textbf{Val Unseen}} \\
		\cmidrule(lr){2-5} \cmidrule(lr){6-9} 
		 & TL~\textdownarrow & NE~\textdownarrow & SR~\textuparrow & SPL~\textuparrow & TL~\textdownarrow & NE~\textdownarrow & SR~\textuparrow & SPL~\textuparrow \\
		\midrule
		DUET~\cite{chen2022think}
		& 12.33 & 2.28 & 79 & 73 
		& 13.94 & 3.31 & 72 & 60 \\
		w/ FSTTA~\cite{gao2024fast}
		& 13.39 & 2.25 & 79 & 73 
		& 14.64 & 3.03 & 75 & 62 \\
		w/ \textsc{ATENA}~(\textbf{Ours})
		& \cellcolor{orange!20}11.27 & \cellcolor{orange!20}2.18 & \cellcolor{orange!20}80 & \cellcolor{orange!20}75
		& \cellcolor{orange!20}12.31& \cellcolor{orange!20}2.90 & \cellcolor{orange!20}75 & \cellcolor{orange!20}66 \\
        \cmidrule(l){1-9}
		BEVBert~\cite{an2023bevbert}
		& 13.56 & 2.17 & 81 & 74
		& 14.55 & 2.81 & 75 & 64\\
		w/ FSTTA$^{\dagger}$
		& 12.28 & 2.31 & 80 & 75 
		& 13.96 & 2.89 & 74 & 63 \\
		w/ \textsc{ATENA}~(\textbf{Ours})
		& \cellcolor{orange!20}10.79 & \cellcolor{orange!20} 2.26 & \cellcolor{orange!20}82 & \cellcolor{orange!20}78
		& \cellcolor{orange!20}12.22 & \cellcolor{orange!20}2.78 & \cellcolor{orange!20}76 & \cellcolor{orange!20}68 \\
		\cmidrule(l){1-9}
		GOAT$^{\dagger}$~\cite{wang2024vision}
		& 11.87 & 1.70 & 84.52 & 79.60
		& 13.43 & 2.33 & 77.91 & 67.34\\
		w/ FSTTA$^{\dagger}$
		& 11.67 & 1.65 & 84.92 & 80.08 
		& 13.26 & 2.32 & 77.99 & 67.48 \\
		w/ \textsc{ATENA}~(\textbf{Ours})
		& \cellcolor{orange!20}11.66 & \cellcolor{orange!20} 1.64 & \cellcolor{orange!20}85.01 & \cellcolor{orange!20}80.13
		& \cellcolor{orange!20}12.52 & \cellcolor{orange!20}2.27 & \cellcolor{orange!20}79.01 & \cellcolor{orange!20}69.30 \\
		\bottomrule
	\end{tabular}}
\end{minipage}
\hfill
\begin{minipage}{0.5\linewidth}
\caption{Experimental results on the R2R-CE dataset.}
\vspace{-0.2em}
\label{tab:r2rce}
\resizebox{\linewidth}{!}{
    \renewcommand{\arraystretch}{1.36}
    {\Huge
    \begin{tabular}{l|ccccc|ccccc}
        \toprule
        \multicolumn{1}{c|}{\multirow{2}{*}{\textbf{Methods}}}
        & \multicolumn{5}{c|}{\textbf{Val Seen}}
        & \multicolumn{5}{c}{\textbf{Val Unseen}} \\
        \cmidrule(lr){2-6} \cmidrule(lr){7-11}
        & TL $\downarrow$ & NE $\downarrow$ & OSR $\uparrow$ & SR $\uparrow$ & SPL $\uparrow$
        & TL $\downarrow$ & NE $\downarrow$ & OSR $\uparrow$ & SR $\uparrow$ & SPL $\uparrow$ \\
        \midrule
        ETPNav~\cite{an2024etpnav}
        & 11.78 & 3.95 & 72 & 66 & 59
        & 11.99 & 4.71 & 65 & 57 & 49 \\
        w/ FSTTA$^{\dagger}$~\cite{gao2024fast}
        & 11.35 & 3.93 & 72 & 66 & 59
        & 11.57 & 4.77 & 64 & 57 & 49 \\
         w/ \textsc{ATENA}~(\textbf{Ours})
        & \cellcolor{orange!20}10.81 & \cellcolor{orange!20}3.86 & \cellcolor{orange!20}72 &\cellcolor{orange!20}67 & \cellcolor{orange!20}61
        & \cellcolor{orange!20}12.89 & \cellcolor{orange!20}4.53 & \cellcolor{orange!20}66 &\cellcolor{orange!20}58 & \cellcolor{orange!20}49 \\
        \cmidrule(l){1-11}
        BEVBert~\cite{an2023bevbert}
        & 13.98 & 3.77 & 73 & 68 & 60
        & 13.27 & 4.57 & 67 & 59 & 50 \\
        w/ FSTTA
        & 14.07 & 4.11 & 74 & 69 & 60
        & 13.11 & 4.39 & 65 & 60 & 51 \\
         w/ \textsc{ATENA}~(\textbf{Ours})
        & \cellcolor{orange!20}11.31 & \cellcolor{orange!20}3.24  & \cellcolor{orange!20}75 &\cellcolor{orange!20}71 & \cellcolor{orange!20}64
        & \cellcolor{orange!20}13.48 & \cellcolor{orange!20}4.50 & \cellcolor{orange!20}67  & \cellcolor{orange!20}60 & \cellcolor{orange!20}51 \\
        \bottomrule
    \end{tabular}
    }
}
\end{minipage}
\vspace{-1em}
\end{table}

\subsection{Main Navigation Results}

\textbf{REVERIE.} Table~\ref{tab:reverie} reports the comparisons of the navigation results on the REVERIE dataset, where the TTA methods including \textsc{ATENA} is applied to HAMT~\cite{chen2021history}, DUET~\cite{chen2022think} and GOAT~\cite{gao2024vision}. Unlike previous methods that utilize entropy minimization as a test-time adaptation signal, we notice a substantial performance increase from \textsc{ATENA}. Specifically, TENT and FSTTA brings minimal performance gains, or rather hinders the navigation performances in several metrics when applied to HAMT and GOAT. However, \textsc{ATENA} improves the SR metric in the validation unseen split up to 3.19\%, 44.98\% and 25.72\% in HAMT, DUET and GOAT, respectively. Furthermore, \textsc{ATENA} also excels in the test unseen split, improving GOAT by 4.59\%, 7.47\%, 15.52\% and 18.13\% in OSR, SR, SPL and RGSPL respectively.

\textbf{R2R \& R2R-CE.} In Table~\ref{tab:r2r}, we present the experimental results on the R2R dataset. Consistent with the findings from the REVERIE dataset, \textsc{ATENA} demonstrates superior effectiveness compared to FSTTA. While FSTTA improves the SPL metric of GOAT by 0.21\% on the validation unseen split, \textsc{ATENA} achieves a 2.91\% improvement. Moreover, for the SR metric on the validation unseen split of DUET, although the success rate is the same as FSTTA, \textsc{ATENA} achieves this with an 11.69\% reduction in trajectory length, highlighting its high navigation efficiency. We observe similar results in the R2R-CE dataset, which is reported in Table~\ref{tab:r2rce}. Specifically, \textsc{ATENA} increases 6.7\% of SPL for BEVBert in the validation seen split. Lastly, we observe that given that the R2R variants rely on dense, step-wise guidance during training, the episodic feedback is relatively sparse to drive significant performance enhancements compared to that of the REVERIE’s.

\begin{table}[t]
\centering
\caption{Comparison of Test-Time Adaptation (TTA) methods with Active Learning (AL). Asterisks~($\ast$) indicate methods integrated with Active Learning~(AL), meaning they receive episodic feedback (success or failure) at uncertain navigation to guide entropy minimization or maximization. Active (\%) denotes the ratio of navigation steps where feedback is requested.}
\label{tab:tta_active_learning}
\renewcommand{\arraystretch}{1.2}
\resizebox{0.95\textwidth}{!}{%
\begin{tabular}{l|cccc|cccc}
\toprule
\multicolumn{1}{c|}{\multirow{2}{*}{\textbf{Methods}}} & \multicolumn{4}{c|}{\textbf{Val Seen}} & \multicolumn{4}{c}{\textbf{Val Unseen}} \\
\cmidrule(lr){2-5} \cmidrule(lr){6-9}
 & SR$\uparrow$ & SPL$\uparrow$ & RGSPL$\uparrow$ & Active (\%)& SR$\uparrow$ & SPL$\uparrow$ & RGSPL$\uparrow$ & Active (\%) \\
\midrule
DUET $+$ TENT$^{\ast}$ & 75.69 & 67.20 & 54.67 & 65.92 & 55.69 & 38.76 & 26.16 & 90.34 \\
DUET $+$ FSTTA$^{\ast}$ & 71.47 & 64.19 & 51.28 & 89.39 & 46.95 & 33.75 & 23.03 & 85.25  \\
\midrule
DUET $+$ MEO$^{\ast}$ (\textbf{Ours}) & \textbf{80.53} & \textbf{72.84} & \textbf{58.78} & \textbf{25.44} & \textbf{63.70} & \textbf{42.49} & \textbf{27.83} & \textbf{55.52} \\
\bottomrule
\end{tabular}}
\vspace{-1em}
\end{table}

\subsection{Comparison of TTA Methods with Active Learning}
Since the compared baseline methods do not employ active learning~(AL) in their framework, we integrate AL into the baselines to highlight the impact of MEO. Specifically, we apply TENT and FSTTA to the pre-trained DUET policy, and allow the agent to update the parameters based on the human evaluation of navigation success or failure at uncertain episodes. Similar to \textsc{ATENA}, these baselines also minimize entropy for successful navigation, and maximize for failed ones. 
Furthermore, we evaluate a variant of \textsc{ATENA} without Self-Active Learning to assess the individual contribution of MEO with AL.
For this experiment, the entropy threshold is equally set as $\delta=0.1$ and we evaluate on the REVERIE dataset. The results are reported in Table~\ref{tab:tta_active_learning}, from which we draw the following observations. First, compared to the result in Table~\ref{tab:reverie}, TENT shows substantial performance increase when guided by human interactions. However, AL provides minimal benefit to FSTTA, which we attribute to its internal mechanism for modifying gradient directions—potentially conflicting with the human-provided guidance on entropy optimization. \textsc{MEO} demonstrates strong synergy with AL, leading to superior navigation performance across SR, SPL, and RGSPL metrics. Moreover, \textsc{MEO} achieves these improvements with significantly fewer human interventions, suggesting that the model progressively gains confidence as navigation proceeds.

\subsection{Effect of Self-Active Learning}
\begin{wraptable}{h!}{0.45\textwidth}
    \vspace{-1.5em}
    \addtolength{\tabcolsep}{-2pt}
    \caption{Comparison of performance on the REVERIE dataset demonstrating the effectiveness of Self-Active Learning~(SAL).}
    \label{tab:self_active_learning}
    \renewcommand{\arraystretch}{1.2}
    \label{tab:pseudo_gt}
    \centering
    \resizebox{\linewidth}{!}{%
\begin{tabular}{l|ccc|ccc}
\toprule
\multicolumn{1}{c|}{\multirow{2}{*}{\textbf{Methods}}}  & \multicolumn{3}{c|}{\textbf{Val Seen}} & \multicolumn{3}{c}{\textbf{Val Unseen}} \\
\cmidrule(lr){2-4} \cmidrule(lr){5-7}
 & SR$\uparrow$ & SPL$\uparrow$ & RGSPL$\uparrow$ & SR$\uparrow$ & SPL$\uparrow$ & RGSPL$\uparrow$ \\
\midrule
w/o SAL & 80.53  &  72.84  &  58.78  &   63.70  &  42.49  &  27.83 \\
w/ SAL & \textbf{84.33} & \textbf{74.31} & \textbf{59.99} & \textbf{68.11} & \textbf{45.82} & \textbf{31.26} \\
\bottomrule
\end{tabular}}
    \vspace{-0.5em}
\end{wraptable}
Table~\ref{tab:self_active_learning} demonstrates the effectiveness of our propose Self-Active Learning~(SAL). We compare the performance of \textsc{ATENA} with and without SAL, applied to the DUET navigation policy. The evaluation is done using the REVERIE dataset. Even without SAL, as previously observed in Table~\ref{tab:tta_active_learning}, \textsc{ATENA} shows a solid performance increase from the base policy.
However, enabling the agents to train online from episodic labels and autonomously evaluate its navigation outcome brings notable improvements. Specifically, we observe in the validation unseen split 6.92\%, 7.84\% and 12.32\% enhancements in SR, SPL, and RGSPL, respectively. This clearly demonstrates the effectiveness of remaining continuously active throughout the adaptation process.

\subsection{Combination Weight of Mixture Entropy Optimization}

\begin{figure}[t]
\begin{center}
\includegraphics[width=\linewidth]{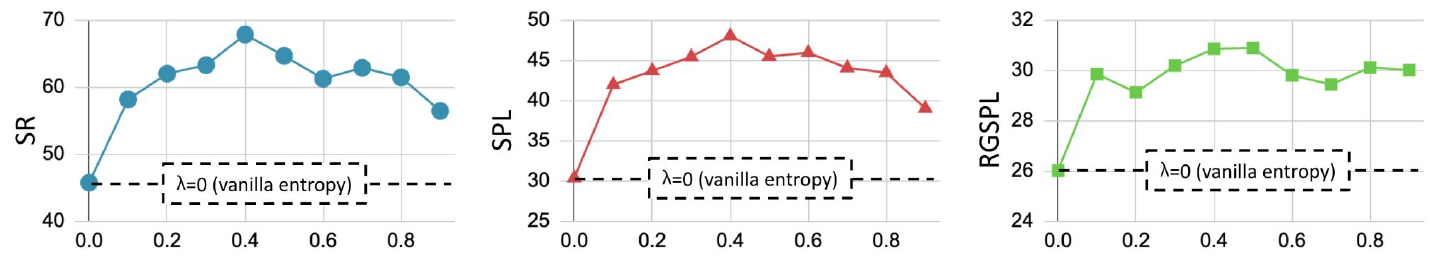}
\end{center}
\vspace{-1.0em}
\caption{\textbf{Effect of the combination weight $\lambda$. } Performance comparison with different combination weight $\lambda$ in Mixture Entropy Optimization. $\lambda=0$ corresponds to vanilla entropy without distribution mix. The results are averaged across three experiments with different seeds.}
\label{fig:lambda}
\vspace{-1.5em}
\end{figure}

We vary the combination weight $\lambda$ in Eq.~\ref{eq:mix_action} within $\lambda \in \{0.0, 0.1, 0.2, \dots, 1.0\}$ and evaluate its effect on adaptation performance. $\lambda = 0.0$ implies that the agent relies solely on its original action distribution, serving as the baseline in Figure~\ref{fig:lambda}. We omit $\lambda = 1.0$ from our results because, in this case, the policy relies exclusively on the entropy of the pseudo-expert distribution, which is identically zero and thus provides no informative signal. As the weight of the pseudo-expert distribution increases, we observe consistent improvements across SR, SPL, and RGSPL, demonstrating the benefit of distribution sharpening. The performance peaks at $\lambda = 0.4$, where the balance between the predicted distribution and the pseudo-expert distribution appears to be optimal. As $\lambda$ increases further, we observe decrease in SR and SPL, yet the benefits compared to vanilla entropy remains solid. These results collectively suggest that while the optimal performance depends on a careful balance between the predicted and pseudo-expert distributions, incorporating the mixture itself is consistently beneficial for adaptation.

\vspace{-0.15em}
\subsection{Sampling Strategies for Feedback Episodes}
\noindent
\begin{wrapfigure}{r}{0.55\textwidth}
  \centering
  \vspace{-1.em}
  \resizebox{\linewidth}{!}{%
      \includegraphics{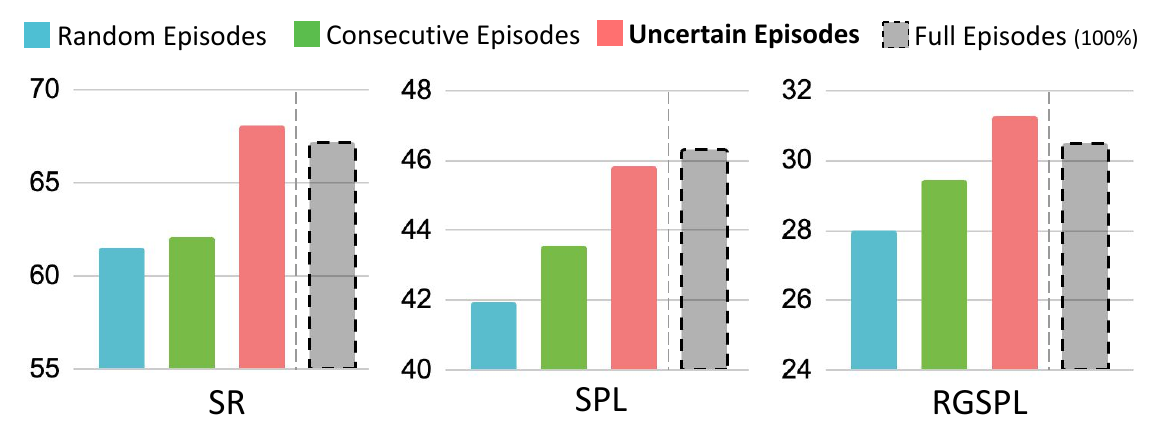}
    }
  \vspace{-1.5em}
  \caption{\textbf{Different Episode Sampling Strategies. }Our uncertainty-based sampling outperforms baselines and remains competitive against full-feedback settings.}
  \vspace{-1.0em}
  \label{fig:sample}
\end{wrapfigure}
\noindent
We compare the uncertainty-based active learning strategy with two different sampling baselines: (1)~Random Episodes, where episodes that receive feedback are selected randomly; and (2)~Consecutive Episodes, where feedback is provided on a contiguous block of episodes starting from the beginning of the dataset. The number of samples of the baselines are equally set to match that of our method's—60\% in this experiment—as the portion of uncertainty-based selection cannot be approximated heuristically. In Figure~\ref{fig:sample}, our uncertainty-based sampling outperforms the baselines, indicating that optimization guided by informative uncertainty signals is more effective than relying on simple rule-based selection. Furthermore, when compared to the setting where feedback is provided for all episodes, our method achieves higher performance in SR and RGSPL, while maintaining competitive results in SPL. These results suggest that our uncertainty-based strategy achieves superior performance with reduced supervision by selectively focusing on the most informative episodes, outperforming both heuristic baselines and full feedback in terms of efficiency and effectiveness.

\vspace{-0.15em}

%% file: sec/6_conclusion.tex
\vspace{-0.5em}
\section{Conclusion}
\vspace{-0.5em}
We introduce \textsc{ATENA}, a novel TTA framework that leverages active human-robot interaction to enhance online vision-language navigation. Specifically, we propose Mixture Entropy Optimization, explicitly reinforcing correct actions and penalizing incorrect ones based on episodic outcomes. Additionally, through Self-Active Learning, we enable the agent to autonomously predict navigation outcomes during episodes where it has relatively high confidence. Extensive experiments demonstrate that \textsc{ATENA} substantially outperforms baseline approaches, effectively addressing distribution shifts between training and testing environments. By integrating human-guided and self-guided active learning mechanisms, \textsc{ATENA} allows the agent to handle uncertainty through continuous adaptation and self-refinement. Ultimately, our approach opens promising avenues for future research by integrating human-robot interaction with automated self-assessment to support robust and efficient online adaptation across diverse interactive embodied AI tasks. \\
\noindent \textbf{Limitations and Future Work.}
A potential limitation of our approach is the generalizability to navigation tasks beyond VLN. Our method specifically leverages active learning through human-robot interaction, naturally aligning with the interactive nature of the VLN task. Therefore, the benefits of our method on less human-interactive navigation tasks, such as Visual Navigation or Object-Goal Navigation, remains underexplored. Investigating the effectiveness of \textsc{ATENA} across diverse navigation benchmarks and additional baselines is necessary to thoroughly assess its broader applicability and robustness.

%% file: sec/7_appendix.tex
\appendix

\section*{\Large Appendix}


\subsection{Details of Evaluation Metrics}

In this section, we provide detailed explanations the metrics employed for evaluating navigation performance in our experiments:

\begin{itemize}[leftmargin=*]

\item \textbf{Trajectory Length (TL):} The average distance traveled by the agent during navigation, measured in meters. A shorter TL indicates more efficient navigation independent of success.

\item \textbf{Navigation Error (NE)} \textit{(R2R only)}: The average shortest-path distance from the agent's final position to the target, measured in meters. A lower NE indicates better localization accuracy.

\item \textbf{Success Rate (SR):} The percentage of episodes in which the agent stops within a threshold distance (typically 3 meters) from the target location. A higher SR indicates better navigation accuracy.

\item \textbf{Oracle Success Rate (OSR):} The percentage of episodes in which the agent passes within the success threshold at any point during navigation. A higher OSR indicates better navigation potential assuming optimal stopping.

\item \textbf{Success weighted by Path Length (SPL):} The average success rate weighted by path efficiency, defined as \( \text{SPL} = \frac{1}{N}\sum_{i=1}^{N} S_i \frac{L_i^*}{max(L_i,L_i^*)} \), where $S_i\in\{0,1\}$ is the success indicator, $L_i^*$ is the shortest path length, and $L_i$ is the actual path length for episode $i$. A higher SPL indicates more efficient and accurate navigation.

\item \textbf{Remote Grounding Success (RGS)} \textit{(REVERIE only)}: The percentage of episodes in which the agent successfully identifies the target object upon stopping, determined by a bounding box IoU of at least 50\%. A higher RGS indicates better navigation and object grounding accuracy.

\item \textbf{Remote Grounding SPL (RGSPL)} \textit{(REVERIE only)}: A metric integrating RGS with path efficiency, similarly penalizing successful grounding by trajectory length as SPL does. A higher RGSPL indicates efficient navigation combined with accurate object grounding.

\end{itemize}

\subsection{Implementation Details}
We search the uncertainty threshold $\delta$ within \{0.1, 0.2, 0.3\}, and the mixture weight $\lambda$ from 0 to 1 at 0.1 intervals. Learning rates are chosen from \{5e-6, 1e-6, 5e-7\}. The rest of the experimental configurations strictly follow the ones of the pre-trained navigation policies we used for the experiment. To simulate the real-world online test-time adaptation scenarios, we use a batch size of 1. All experiments are conducted on a single NVIDIA RTX 3090 GPU, though the method is lightweight enough to run on lower-powered on-device hardware. The results in the experiment are averaged over 3 different random seeds. To ensure a fair comparison with the episodic update nature of \textsc{ATENA}, we implement TENT~\cite{wang2020tent} in VLN baselines by performing parameter updates at the end of each episode. For FSTTA~\cite{gao2024fast}, we set the intervals for slow and fast updates to 4 and 3, respectively, following the original work. However, we modify the learning rates, selecting the fast learning rate from \{6e-3, 6e-4, 6e-5\} and the slow learning rate from \{5e-3, 1e-3, 3e-4\}, due to the lack of reproducibility in the original codebase~\footnote{\url{https://github.com/Feliciaxyao/ICML2024-FSTTA/issues/1}}, aiming to approximate the reported results as closely as possible.

\subsection{Leaderboard Results}

Table~\ref{tab:leaderboard} shows the REVERIE challenge leaderboard ranks~\footnote{\url{https://eval.ai/web/challenges/challenge-page/606/leaderboard/1683}} on the test-unseen split, ordered by Success Rate (SR). When integrated with GOAT~\cite{wang2024vision}, our proposed method, \textsc{ATENA}, achieves third place on the official leaderboard as of submission date. \textsc{ATENA} offers competitive performance through a lightweight, easily integrable approach that does not necessitate additional pretraining or substantial structural changes, highlighting its practical effectiveness and compatibility with advanced models. In contrast, the top-ranked models, RREx-BoT and RREx-BoT Pre-Explore~\cite{sigurdssonrrex}, rely heavily on extensive pretraining with a large-scale vision-language architecture and multiple sophisticated data augmentation techniques. Despite their higher SR and OSR scores, \textsc{ATENA} surpasses RREx-BoT in terms of SPL and RGSPL metrics, further demonstrating its efficiency and practicality in real-world navigation tasks.

\begin{table}[htbp]
\centering
\caption{Leaderboard Performance Comparison}
\resizebox{0.7\columnwidth}{!}{%
\begin{tabular}{clcccc}
\toprule
Rank & Model & SR~$\uparrow$ & OSR~$\uparrow$ & SPL~$\uparrow$ & RGSPL~$\uparrow$ \\
\midrule
1 & RREx-BoT Pre-Explore~\cite{sigurdssonrrex} & 65.19 & 73.74 & 62.04 & 40.12 \\
2 & RREx-BoT~\cite{sigurdssonrrex} & 65.18 & 100.00 & 42.07 & 2.78 \\
\cellcolor{orange!20}{\textbf{3}} & \cellcolor{orange!20}{GOAT~\cite{wang2024vision} w/ \textbf{ATENA (ours)}} & \cellcolor{orange!20}{62.03} & \cellcolor{orange!20}{64.26} & \cellcolor{orange!20}{46.82} & \cellcolor{orange!20}{31.54} \\
4 & SRVLN & 61.38 & 66.16 & 35.62 & 27.12 \\
5 & VinciG & 60.46 & 64.86 & 41.33 & 27.84 \\
\bottomrule
\end{tabular}
}
\label{tab:leaderboard}
\end{table}

\subsection{Precision of Self-Prediction Head}
\definecolor{good}{RGB}{204,238,204}  
Since agents rely on Self-Prediction head for relatively certain navigation episodes, its reliability is crucial for stable adaptation. For instance, if Self-Prediction head were unreliable, adaptation could collapse due to false signals. 
To empirically assess the reliability of our Self-Prediction head, we analyze its predictions of the navigation outcome on the validation unseen split of the REVERIE dataset. The confusion matrix shown in Table~\ref{tab:confusion-matrix} demonstrates high reliability with 541 out of 757 negatives (71.46\%) and notably 908 out of 1006 positives (90.26\%) predicted correctly.
\begin{wraptable}{h!}{0.375\textwidth}
    \vspace{-1.5em}
    \addtolength{\tabcolsep}{-2pt}
    \caption{Confusion matrix of Self-Prediction head}
    \renewcommand{\arraystretch}{1.2}
    \label{tab:confusion-matrix}
    \centering
    \resizebox{\linewidth}{!}{%
    \begin{tabular}{cl|cc}
\toprule
 & & \multicolumn{2}{c}{\textbf{Prediction}} \\[-2pt]  
\cmidrule(lr){3-4}
 & & Positive & Negative \\[-2pt]
\midrule
\multirow{2}{*}{\rotatebox[origin=c]{90}{\textbf{Actual}}}
 & Positive & \cellcolor{good}\textbf{908} & 98  \\
 & Negative & 216 & \cellcolor{good}\textbf{541} \\ 
\bottomrule
\end{tabular}}
    \vspace{-0.5em}
\end{wraptable}
Its overall prediction accuracy of 82.19\% indicates robust reliability, making it suitable as a pseudo-label source for self-supervised signal. While predictions are not entirely perfect, the proportion of incorrect predictions is significantly lower compared to correct ones, and potential inaccuracies can be mitigated by reducing the weight of self-predicted outputs by $\gamma$ in Eq.~\ref{eq:total_loss}. Consequently, as discussed in Table~\ref{tab:self_active_learning} of the main paper, the proposed Self-Prediction head effectively contributes to performance improvements.

\subsection{Computational Cost}

In this section, we analyze the computational cost of \textsc{ATENA}.
In Table~\ref{tab:compute_cost_ms}, Nav. time refers to the average duration~(ms) taken by the agent to perform a navigation rollout, and Adapt. time denotes the average duration~(ms) spent updating the policy in between the episodes.
For DUET~\cite{chen2022think} we only measures the navigation time without performing any adaptation. FSTTA~\cite{gao2024fast} continuously updates its policy during navigation rollout, significantly increasing navigation latency(141.55 ms → 1,155.86 ms), whereas TENT~\cite{wang2020tent} and our ATENA introduce minimal additional latency during navigation, which is due to additional collection and calculation of entropy. Although ATENA slightly increases adaptation latency (+0.93\%) compared to that of TENT's, this minor increase is justified by ATENA’s substantial performance improvements over TENT (SR +20.56\%, OSR +20.45\%). While ATENA's adaptation time exceeds that of FSTTA, latency during navigation is more detrimental in real-world robotic tasks. Thus, compared to other existing TTA methods in VLN, ATENA stands out as the most efficient and practical approach.

\begin{table}[ht!]
\centering
\caption{Comparison of the average computation time per episode~(ms) for TTA methods}
\label{tab:compute_cost_ms}
\renewcommand{\arraystretch}{1.2}
\resizebox{0.8\columnwidth}{!}{%
\begin{tabular}{l|cccc|c|c|c}
\toprule
\multicolumn{1}{c|}{\multirow{2}{*}{\textbf{Methods}}} & \multicolumn{6}{c}{\textbf{REVERIE - Val Unseen}}\\
\cmidrule(lr){2-8}
 & OSR~$\uparrow$ & SR~$\uparrow$ & SPL~$\uparrow$ & RGSPL~$\uparrow$ & Nav. Time (ms) & Adapt. Time (ms) & Total (ms)\\
\midrule
DUET~\cite{chen2022think}                     & 51.07 & 46.98 & 33.73 & 23.03 & 141.55 & -- & 141.55 \\
w/ TENT~\cite{wang2020tent}                   & 51.43 & 47.55 & 33.99 & 23.32 & 156.23 & 244.68 & 400.91 \\
w/ FSTTA~\cite{gao2024fast}                   & 56.26 & 54.15 & 36.41 & 23.56 & 1,155.86 & 45.78  & 1,201.64 \\
\cellcolor{orange!20}{w/ ATENA (\textbf{Ours})} & \cellcolor{orange!20}{71.88} & \cellcolor{orange!20}{68.11} & \cellcolor{orange!20}{45.82} & \cellcolor{orange!20}{32.26} & \cellcolor{orange!20}{163.73} & \cellcolor{orange!20}{246.96} & \cellcolor{orange!20}{410.69} \\
\bottomrule
\end{tabular}}
\end{table}

\subsection{Broader Impact}
With increasing emphasis on enhancing human-robot interactions, developing effective methods to facilitate these interactions has become crucial. In line with this trend, our proposed method, \textsc{ATENA}, provides a novel approach enabling robots to adapt effectively to dynamic and complex environments based on individual user feedback. Consequently, \textsc{ATENA} contributes to improved user-centric performance, reliability, and robustness in real-world applications involving active human interaction. However, ambiguity or inconsistency in user feedback might introduce errors in the system's interpretation, potentially diminishing overall performance and user satisfaction. Therefore, further research into accurately interpreting ambiguous user feedback remains essential.

\subsection{Trajectory Visualizations}
\begin{figure}[htbp]
    \centering
    \caption{
    In Step 3, Trial 1 selected an incorrect action with 94.5\% confidence. After ATENA's adaptation, Trial 2 correctly choose the appropriate action with 99.9\% confidence. The intended target—the faucet in the bathroom with a \textbf{dark green hand towel}—was missed in Trial 1 (final distance: 10.56), but successfully located in Trial 2.}
    \vspace{+1.em}
    \includegraphics[width=\textwidth]{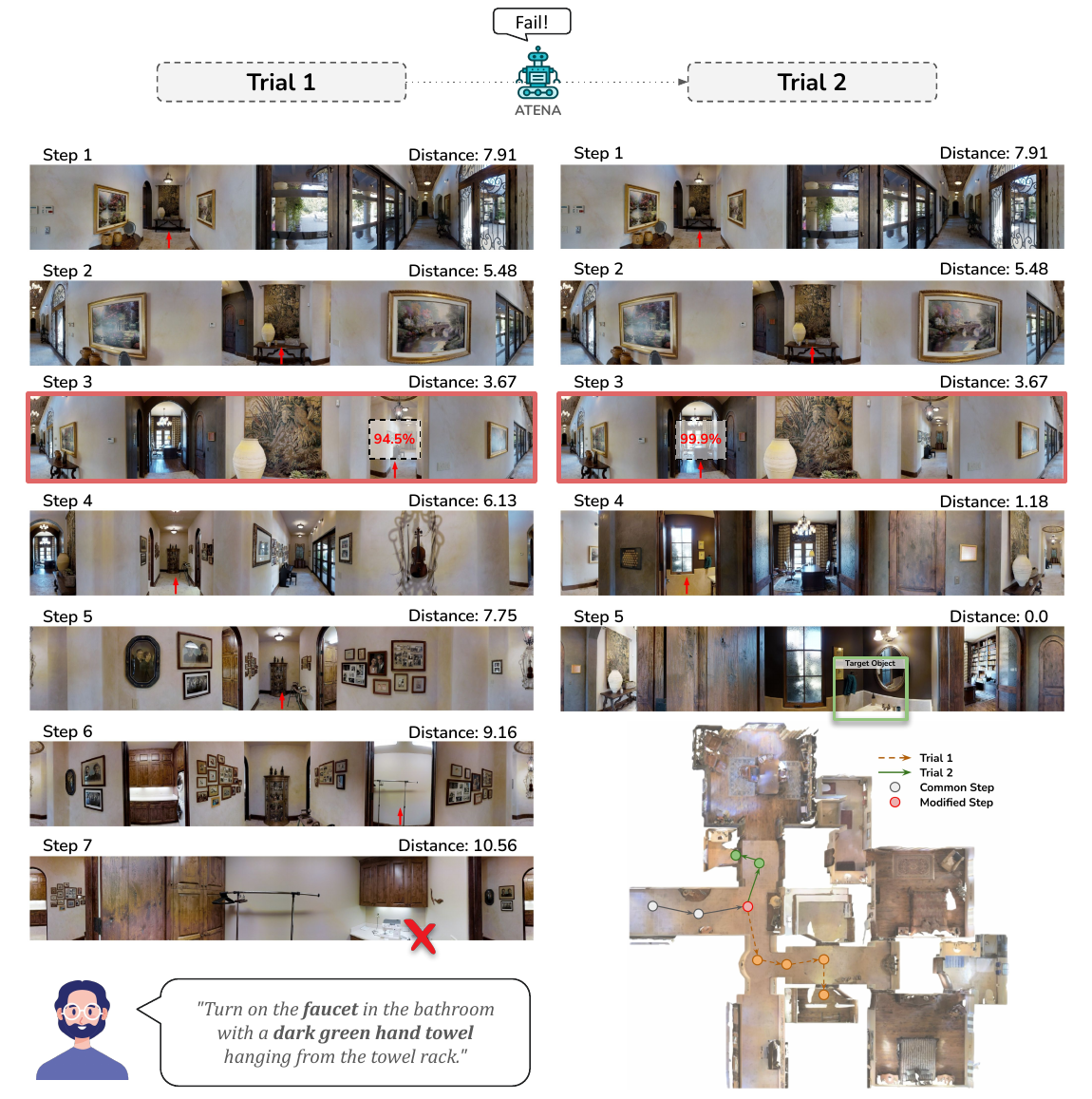} 
    \label{fig:pano1}
\end{figure}

\begin{figure}[htbp]
    \centering
    \caption{
    In Step 2, Trial 1 incorrectly selected an action with 85.4\% confidence. After ATENA's adaptation, Trial 2 correctly selects the action with 92.7\% confidence. The target object—the \textbf{blue pillow} on the beige sofa—was missed in Trial 1 (final distance: 10.69) but was successfully located in Trial 2.}
    \vspace{+1.em}
    \includegraphics[width=\textwidth]{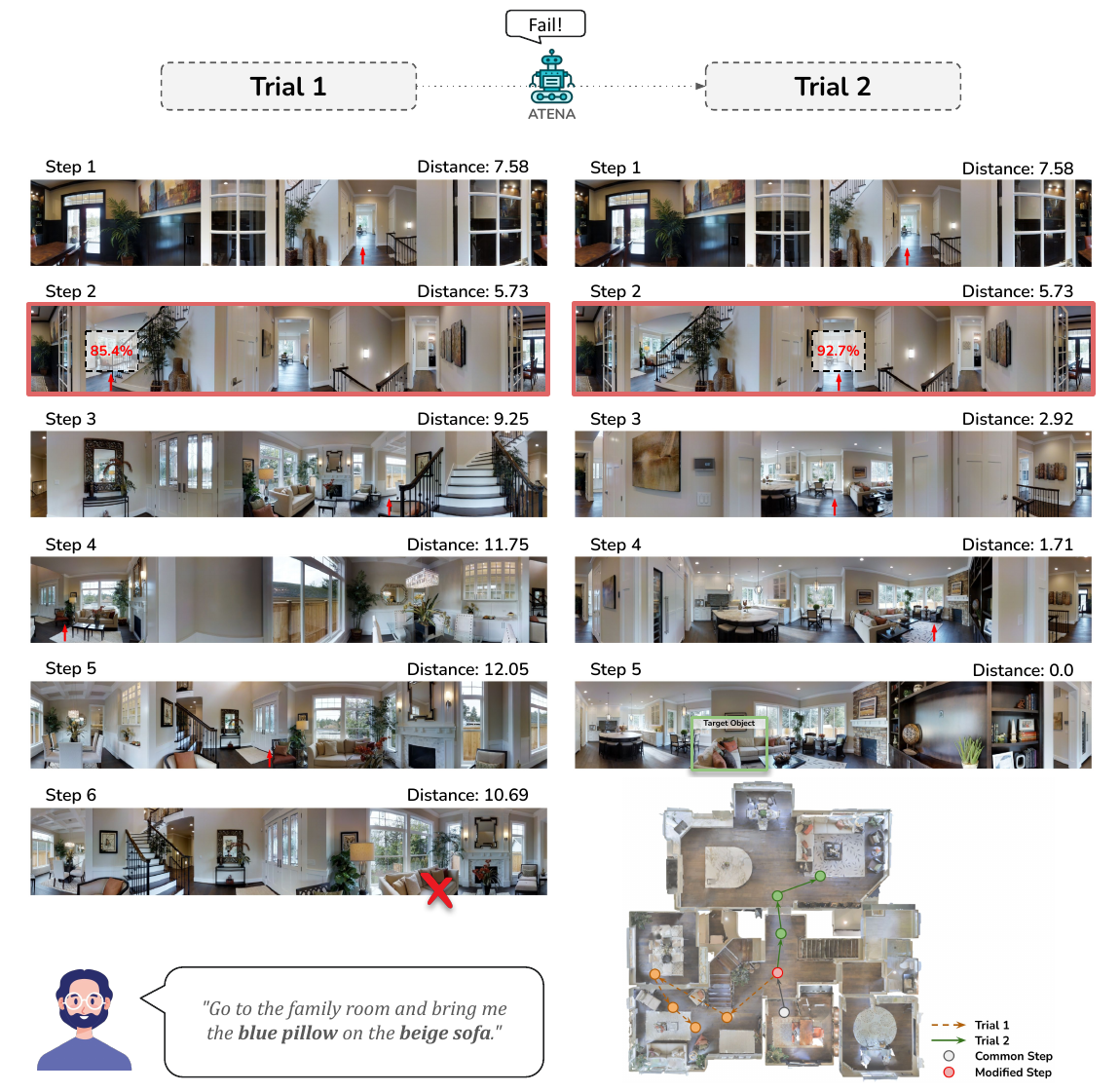} 
    \label{fig:pano2}
\end{figure}

\begin{figure}[htbp]
    \centering
    \caption{
    In Step 3, Trial 1 incorrectly selected an action with 98.1\% confidence. After adaptation with ATENA, Trial 2 correctly identifies the target action with 100\% confidence. The intended target, the \textbf{plant on the bathroom counters}, was mistakenly identified in Trial 1 as the plant above the toilet (final distance: 8.3) but successfully located in Trial 2.}
    \vspace{+1.em}
    \includegraphics[width=\textwidth]{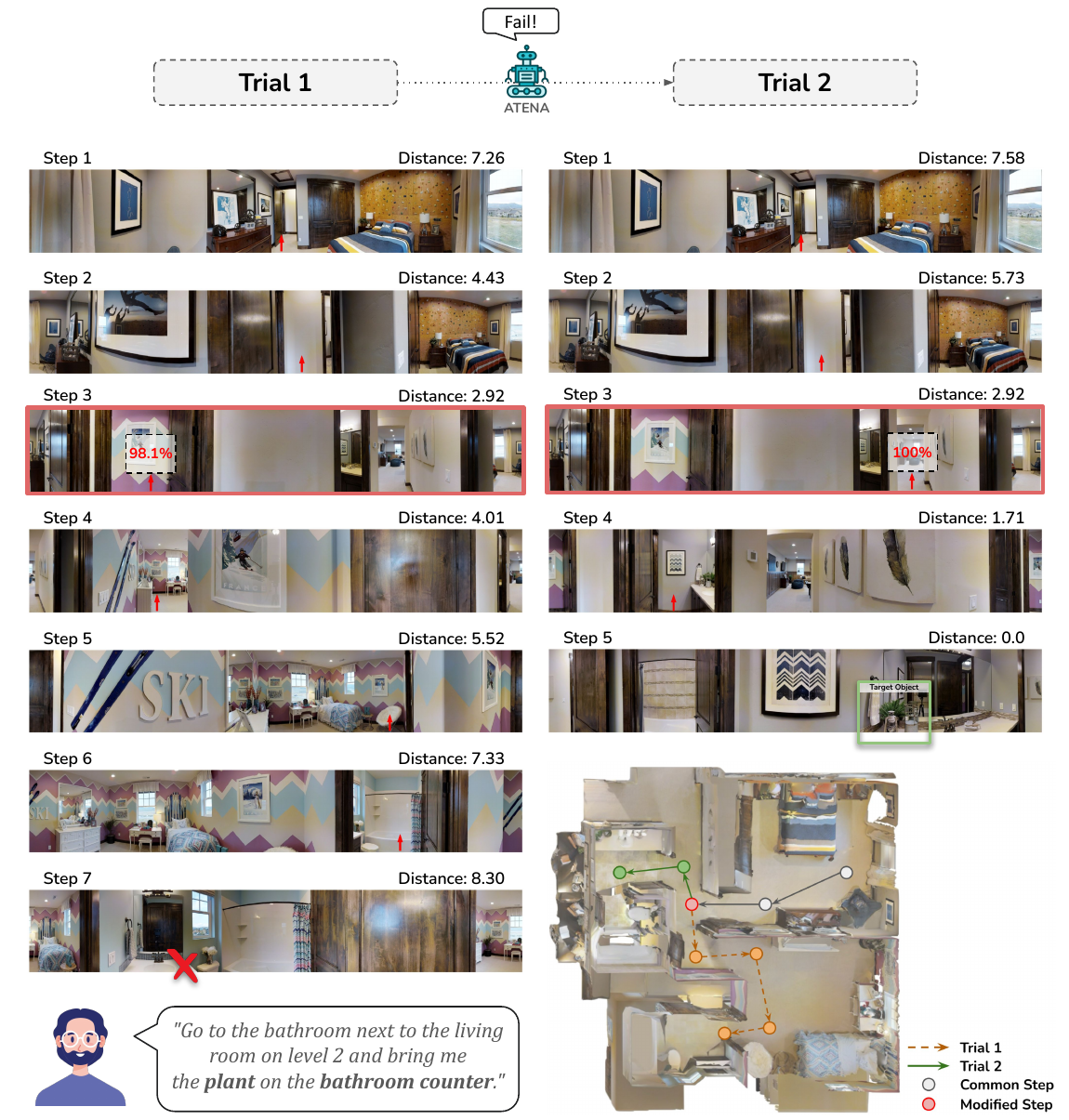} 
    \label{fig:pano3}
\end{figure}

We illustrates the trajectories of ATENA integrated with DUET in Figure~\ref{fig:pano1}, ~\ref{fig:pano2}, \ref{fig:pano3}, conducted using the REVERIE~\cite{qi2020reverie} dataset. In the figure, Trial 1 shows the trajectory before adaptation, where it fails to reach the target destination in the first trial. Trial 2 shows the trajectory after adaptation with our ATENA, which successfully reaches the target destination. The red boxes, defined as Modified Step, highlight navigation points with a high probability of incorrect navigation actions in Trial 1, which shift to a high probability of correct actions in Trial 2 after adaptation, which as a result contributes to success. Green boxes denote that the target object was successfully found, and distances indicate proximity to the target.

\subsection{Code base and License}

Table~\ref{tab:license_terms} provides detailed information about the licenses and official URLs for the datasets and simulators utilized throughout our experiments.

\begin{table}[htbp]
\centering
\caption{License and URLs for Datasets and Simulators}
\resizebox{0.95\columnwidth}{!}{%
\renewcommand{\arraystretch}{1.1}
\large
\begin{tabular}{c c c}
\toprule
\textbf{Name} & \textbf{License} & \textbf{URL} \\
\midrule
R2R~\cite{anderson2018vision} & Matterport3D Terms of Use & \url{https://bringmeaspoon.org} \\
REVERIE~\cite{qi2020reverie} & Matterport3D Terms of Use & \url{https://github.com/YuankaiQi/REVERIE} \\
R2R-CE~\cite{krantz2020beyond} & Matterport3D Terms of Use & \url{https://github.com/jacobkrantz/VLN-CE} \\
Matterport3D~\cite{chang2017matterport3dlearningrgbddata} & Matterport3D Terms of Use & \url{https://niessner.github.io/Matterport/} \\
Habitat Simulator~\cite{habitat19iccv} & MIT License & \url{https://github.com/facebookresearch/habitat-sim} \\
\bottomrule
\end{tabular}}
\label{tab:license_terms}
\end{table}

\newgeometry{left=2cm,right=2cm}
\restoregeometry

\newgeometry{left=2cm,right=2cm}
\restoregeometry

\newgeometry{left=2cm,right=2cm}
\restoregeometry